\documentclass[11pt,a4paper]{article}
\usepackage[hyperref]{eacl2021}
\usepackage{times}
\usepackage{latexsym}
\usepackage{soul}
\usepackage{graphicx}
\usepackage{subcaption}

\usepackage{microtype}

\aclfinalcopy 

\title{WASSA@IITK at WASSA 2021: Multi-task Learning and Transformer Finetuning for Emotion Classification and Empathy Prediction}

\author{
    Jay Mundra$^{*}$ \qquad   
    Rohan Gupta$^{*}$ \qquad 
    Sagnik Mukherjee\thanks{\quad Authors equally contributed  to this work.} \qquad  \\
{Indian Institute of Technology Kanpur (IIT Kanpur)} \\
{\tt \{jaym, rohangpt, sagnikm\}@iitk.ac.in}\\
}

\date{}

\begin{document}
\maketitle
\begin{abstract}
This paper describes our contribution to the WASSA 2021 shared task on Empathy Prediction and Emotion Classification. The broad goal of this task was to model an empathy score, a distress score and the overall level of emotion of an essay written in response to a newspaper article associated with harm to someone. We have used the ELECTRA model abundantly and also advanced deep learning approaches like multi-task learning. Additionally, we also leveraged standard machine learning techniques like ensembling. Our system achieves a Pearson Correlation Coefficient of \textbf{0.533} on sub-task I and a macro F1 score of \textbf{0.5528} on sub-task II. We ranked \textbf{1$^{st}$} in Emotion Classification sub-task and \textbf{3$^{rd}$} in Empathy Prediction sub-task.
\end{abstract}

\section{Introduction}
With the growing interest in human-computer interface, machines still lag in having and understanding emotions. Emotions are considered a trait of a living being and are often used to list differences between machines and living beings. Human emotions such as empathy are hard to describe even for humans, and a consensus is hard to be reached. The inherent knowledge humans have for these emotions is hard to pass on to machines, and hence this task is challenging. That is also probably the reason why the prior work in this area is really limited. Although there has been some research done by \citet{6411762}, \citet{Gibson2015PredictingTE} and \citet{khanpour-etal-2017-identifying}, they have some significant limitations, such as the oversimplified notion of empathy and unavailability of these corpora in the public domain.  \citet{sedoc2020learning} investigated the utility of Mixed-Level Feed Forward Network and also created an empathy lexicon.\\
Another problem in the affect domain is that of emotion classification of textual data. \citet{klinger2018analysis} presented an analysis of many datasets for this task. These datasets vary in size, origin and taxonomy. The most frequent emotion taxonomy used is that proposed by \citet{doi:10.1080/02699939208411068} - identifying anger, disgust, fear, joy, sadness and surprise as the 6 emotion categories. \citet{demszky2020goemotions} presented a dataset obtained from Reddit comments and tagged according to 27 fine-grained emotion categories identified by them. They also present a baseline obtained by fine-tuning a BERT-base \cite{devlin2019bert} model on their dataset.\\
In the WASSA 2021 Shared Task \cite{tafreshi-etal-2021} there are two major sub-tasks addressing the issues of empathy and distress prediction and emotion classification.\\\textbf{Track I: Empathy Prediction - } the objective is to model the empathy concern as well as the personal distress at the essay level. This is a regression task. \\
\textbf{Track II: Emotion Prediction - } the objective is to predict the overall emotion of the essay. The labels are categorical, and it is a classification task.\\
In our approach, we have used transformer-based language models, primarily ELECTRA \cite{clark2020electra}.
We also experimented across different forms of multi-task learning, keeping ELECTRA at the base and having several feedforward layers on top of it, responsible for different tasks. It was also observed that various ensembling techniques worked pretty well for the tasks and improved the scores due to a better bias-variance trade-off.

\section{Dataset}
For this task, an extended version of the dataset by \citet{buechel-etal-2018-modeling} was used, as per the official release from the task organizers. \\
The dataset contained 1860 data points for training where each data point was a tuple of the following - the essay, an empathy score, a distress score, age and income of the annotator (demographic factors), overall emotion of the essay and various metrics denoting the personality of the respondent. The empathy and distress scores were in the range $(1,7)$, and they were annotated with a 7-point scale. The validation and test sets had 270 and 525 data points, respectively. The emotion classes used in the task are closely related to the emotion classes discussed in \citet{doi:10.1080/02699939208411068}.\\
For this task, the volume of data was insufficient for finetuning large language models like ELECTRA. To overcome this, we experimented with data augmentation; in particular, we used the Ekman grouped data for emotion classification provided in the GoEmotions dataset \cite{demszky2020goemotions}. The pros and cons of this methodology will be discussed in Section \ref{sys_des:data_au}\\

\section{System Description}
\label{sec:length}
\subsection{Empathy Prediction}
We propose two separate approaches for this task. 
\begin{figure}
    \centering
    \includegraphics[height=4cm ,width=7 cm]{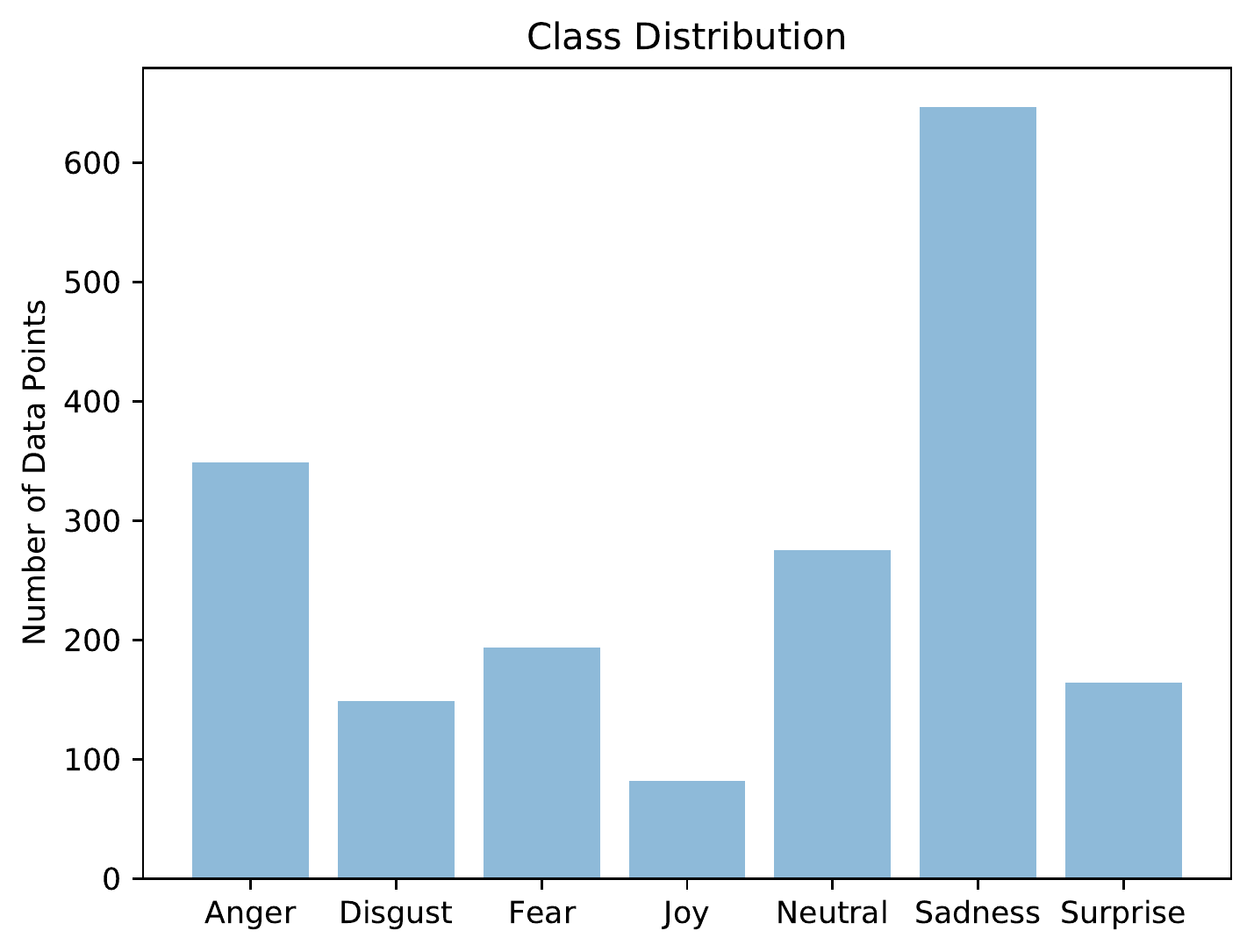}
    \caption{Frequency of classes in the dataset} 
    \label{fig:x}
\end{figure}
\subsubsection{Transformer finetuning}
\label{sys_des:vanilla}
The first approach is based on finetuning an ELECTRA large model separately for empathy and distress with the Mean Squared Error (MSE) loss function. The ELECTRA parameters were kept trainable, and it was finetuned with a single feed-forward layer on top of it. We experimented with the number of linear layers and the number of hidden neurons and chose a single layer since it yielded the best validation score. We shall refer to this approach as the Vanilla ELECTRA method.\\
\subsubsection{Finetuning with Multi-task learning}
\label{sys_des:multitask}
Multi-task learning has recently caught a lot of interest in the NLP community (\citet{liu-etal-2019-multi}, \citet{Worsham_2020}). The objectives, namely empathy and personal distress, are seemingly closely related, and hence a multi-task learning setup was used, jointly modelling them. We used an ELECTRA-large with two dense layers on top of it, one responsible for Empathy and another for Distress. Like the previous approach, MSE loss was used, adding the loss for Empathy and Distress and jointly training the architecture end to end on that total loss.\\
The same approach was tried out with the RoBERTa \cite{liu2019roberta} model.

\subsubsection{Final Ensemble}
\label{sys_des:ensemble}
For the empathy prediction task, the final system was an ensemble of two models - Roberta  multi-task model and Vanilla ELECTRA model.\\
For the distress prediction task, the final system was an ensemble of two models, both being ELECTRA models trained with multi-task loss, with different performances on the development set.
We finally submitted an ensemble of two models for each task - Empathy prediction, and Distress prediction.\\
For all these ensembling, a simple average of the output across the trained models was taken.
\subsection{Emotion Classification}
\subsubsection{Transformer finetuning}
\label{sys_des:emo_sub}
In this approach the ELECTRA model was finetuned. The [CLS] token was passed through a single linear layer to produce a vector of size 7, representing class probabilities (or scores). We trained using the cross entropy loss function.\\
As per our observations these models were sensitive to initialisation. The validation accuracy scores varied significantly across different seed values. Hence, the models were trained several times, and the snapshots with best validation scores were saved.
\subsubsection{Data Augmentation}
\label{sys_des:data_au}
The Figure \ref{fig:x} clearly shows that there was a heavy class imbalance in the dataset provided by the organizers, `joy' being the least represented class (since the data collected is related to harm done to someone). Class imbalance is a standard problem faced by the machine learning community in classification problems (\citet{longadge2013class}, \citet{10.1016/j.eswa.2016.12.035}). To address this issue, the dataset was augmented with the GoEmotions dataset \cite{demszky2020goemotions}, since the class labels of the Ekman variant of it was exactly the same. However, this dataset was different from ours because the essays' length was significantly lesser for GoEmotions dataset than our task dataset. Hence the appropriate number of data points to be sampled was an important hyperparameter. Too much sampling would make the distribution of the train data very different from the distribution of validation and test data, and would cause the model to eventually fail to generalise.\\
The augmentation scheme was different across different components of the final ensemble. While an ELECTRA base was finetuned on a class balanced dataset of total 2800 samples represented by BA (Balanced Augmentation) in Table \ref{acc-emo}, another ELECTRA base was finetuned with randomly chosen 1000 samples  represented by RA (Random Augmentation) in Table \ref{acc-emo}.
Same augmentation scheme was followed on ELECTRA Large and RoBERTa. We also trained an ALBERT \cite{lan2019albert} model on the BA augmented data. The intuition behind using multiple transformer models was to have different ``strength" of different models.
\subsubsection{Final Ensemble}
We created two ensemble models for this task by summing the probability scores of the models involved in ensemble.
The final systems submitted for this task were two ensembles. The ensembling technique was to take sum or average across class scores and then selecting the class with the highest score.
One of them (Ensemble 1 in table \ref{acc-emo}) was an ensemble of two ELECTRA base models and one ELECTRA large model (model 1, 2, 3 in the table \ref{acc-emo}) that were trained on different augmentation schemes.

The second ensemble (Ensemble 2 in table \ref{acc-emo}) is the ensemble of the first 7 models shown in Table \ref{acc-emo} comprising of 2 ELECTRA base, 2 ELECTRA large, 2 RoBERTa, and an ALBERT trained using the methods outlined in the previous subsection.

\begin{table}[t]
\centering
\begin{tabular}{c|c}
\hline
\textbf{Method} & \textbf{Val Macro F1}\\[1 ex]
\hline\hline
+ aug &  0.6042\\
\hline
- aug & 0.561\\
\hline
\end{tabular}
\caption{Variation of validation macro F1 scores with and without the data augmentation technique for classification task for the ELECTRA base model ('+ aug' means data augmentation has been used and '- aug' means otherwise.)}
\label{table:aug_plus_minus}
\end{table}

\begin{figure}
    \centering
    \includegraphics[trim={0.7cm 0 0 0}, clip, height = 5cm, width = 8cm]{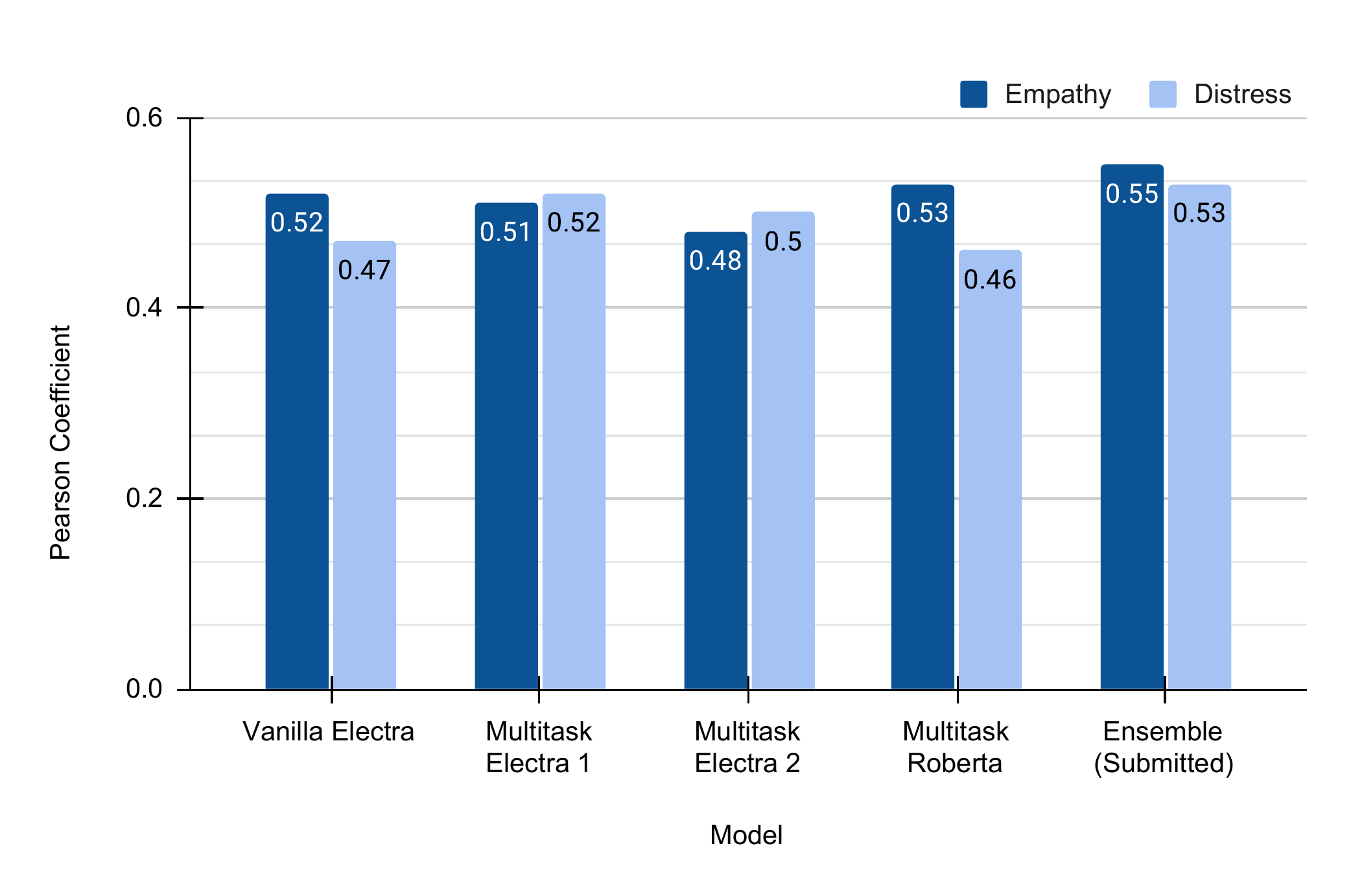}
    \caption{Pearson's Correlation Coefficient of Sub-task I on dev data}
    \label{fig:y}
\end{figure}
\section{Experimental Setup}
For all tasks the learning rate was set to 10$^{-5}$, and the models were trained using AdamW \cite{adam} as optimizer. The parameters for AdamW were $\beta = (0.9, 0.99)$, $\epsilon = 10^{-6}$, weight\_decay = 0. Batch size was set to 16 for the Section \ref{sys_des:vanilla} approach, and set to 8 for Section \ref{sys_des:multitask} and Section \ref{sys_des:emo_sub} with the shuffle parameter set to True on pytorch dataloader. A single Tesla V100-SXM2-16GB GPU was used  for the finetuning experiments. The GPUs were available on the Google Colab platform. \\
The ELECTRA and RoBERTa were used off the shelf from the HuggingFace library \cite{wolf2020huggingfaces}.

\section{Results}
\begin{table}[t]
\centering
\begin{tabular}{l|c|c}
\hline \textbf{Model} & \textbf{Macro F1} & \textbf{Accuracy} \\ \hline\hline
ELECTRA base (RA) & 0.604 & \textbf{69.25} \\
ELECTRA base (BA)& \textbf{0.608} & 67.77\\
ELECTRA Large (RA) & 0.582 & 68.51\\
ELECTRA Large (BA) & 0.585 & 66.29 \\
RoBERTa Large (RA) & 0.588 & 67.03\\
RoBERTa Large (BA) & 0.583 & 66.29 \\
ALBERT Large (BA) & 0.595 & 68.51 \\ \hline

Ensemble 1* & 0.64 & 71.11 \\
Ensemble 2 & \textbf{0.65} & \textbf{72.59} \\ \hline

\end{tabular}
\caption{\label{acc-emo} Accuracy and macro-F1 scores on emotion classification on the dev data. * - Submitted model}
\end{table}

\begin{figure*}[t]
\centering
\begin{subfigure}[b]{.45\linewidth}
\includegraphics[height = 5.5cm, width=\linewidth]{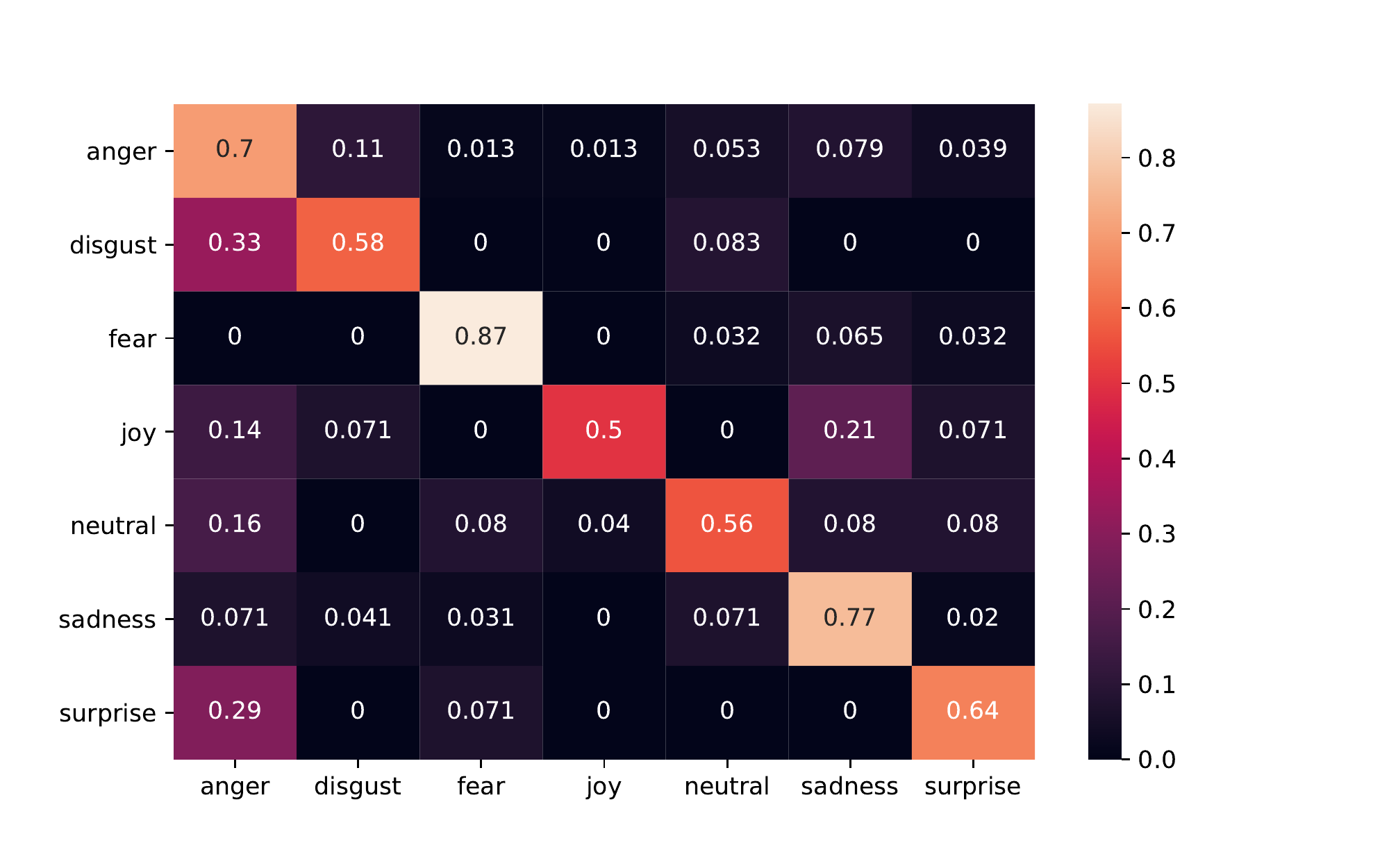}
\caption{Final Submission}\label{fig:mouse}
\end{subfigure}
\begin{subfigure}[b]{.45\linewidth}
\includegraphics[height = 5.5cm, width=\linewidth]{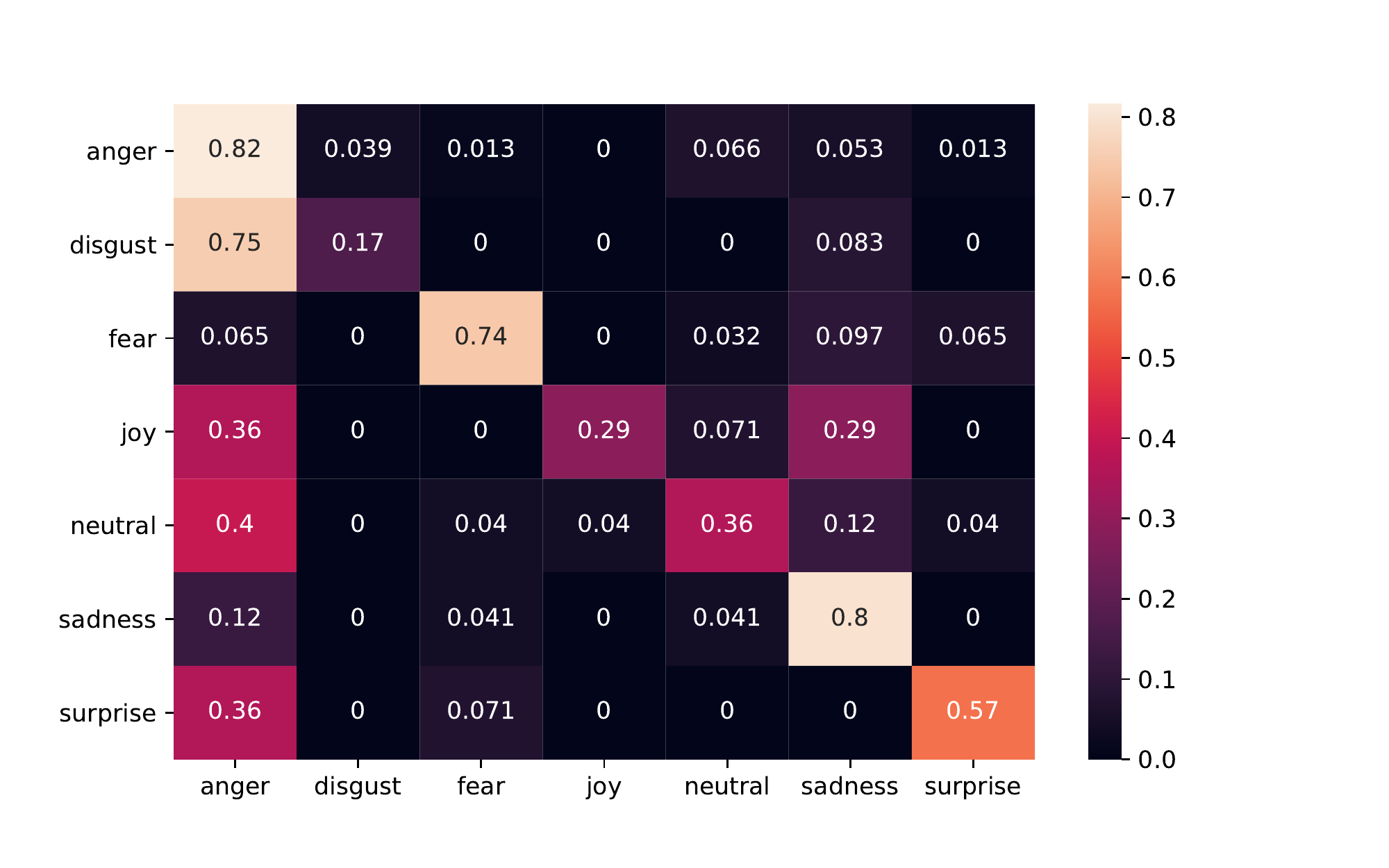}
\caption{Electra base without augmentation}\label{fig:mouse}
\end{subfigure}

\caption{Confusion matrix (normalized) for the emotion classification task on the dev data}
\label{fig:cm}
\end{figure*}

The results on the empathy prediction sub-task and the distress prediction sub-task on the development set can be found in Figure \ref{fig:y}. Also presented in the figure is the performance of the ensemble models that we submitted for evaluation. On the test data, the ensembles achieved a pearson correlation coefficient of \textbf{0.558} and \textbf{0.507} on the Empathy prediction, and Distress prediction respectively. This amounts to an average score of \textbf{0.533} which ranks us 3$^{rd}$ in this sub-task.\\
In Table \ref{table:aug_plus_minus}, we present a result on the development set  with and without using balanced data augmentation (BA) using GoEmotions for the ELECTRA base model. As we expected, the data augmentation helped improve performance of the model. The results for the various EMO models on the development set are available in Table \ref{acc-emo}. We also list the performance of two final ensembles in the table.
On the test data, the Ensemble 1 and Ensemble 2 achieved a macro F1 score of \textbf{0.5528} and \textbf{0.588} respectively. We submitted Ensemble 1 as our final submission in the evaluation phase, as we were allowed only one submission. We ranked 1$^{st}$ in this sub-task.




\section{Error Analysis}
   


It was observed that the training was highly sensitive to the initialization of the models. These include the initializations of the weight vectors of the feed-forward layers, and the ordering and organization of the batches fed to the model. Across different seeds the models' scores varied significantly.  This is in line with the analysis done by \citet{modidodge2020fine} for transformer based models on the GLUE Benchmark \cite{wang2018glue}. The high variability in performance can be observed Fig \ref{fig:y} . The Multi-task ELECTRA model performed differently on different runs, and we list two such runs for each Empathy Prediction and Distress Prediction.\\
The confusion matrix in Figure \ref{fig:cm} shows that while the submitted system performed extremely well on the fear class, it underperformed a bit on joy and neutral, the performance on joy being very low. A point worth noting might be that the only `positive' human emotion here is joy, and all the others are negative emotions and are often hard to distinguish by humans. A fare share of data points have been labeled as sadness and anger while they actually belong to the class joy. We also present, for comparison, the confusion matrix of the ELECTRA base model trained on non-augmented data. This model performed much worse on all emotions except sadness and anger; the model being very prone to predict the emotion as anger. The good performance of this model on sadness could be because of high number of samples from the that class in training data. The tendency to predict anger correlates with the fact that anger is the 2nd most frequent label in the training data. From this comparison, we can say data augmentation helped us achieve a more balanced performance with respect to all emotions, in particular bringing down the tendency to predict anger as the emotion, and improving performance for all other emotion classes which had less training data.

\section{Conclusion}
This paper describes our submission to the WASSA 2021 shared task, where we have leveraged off the shelf transformer models pre-trained on huge corpora in the English language. The intuition for keeping these models was to exploit the huge information they already possess. In the evaluation phase our systems ranked 1st in track II and 3rd in track I.

\bibliography{anthology,eacl2021}
\bibliographystyle{acl_natbib}

\end{document}